\documentclass[Journal,letterpaper]{ascelike-new}
\usepackage[utf8]{inputenc}
\usepackage[T1]{fontenc}
\usepackage{lmodern}
\usepackage{graphicx}
\usepackage[figurename=Fig.,labelfont=bf,labelsep=period]{caption}
\usepackage{subcaption}
\usepackage{amsmath}
\usepackage{newtxtext,newtxmath}
\usepackage[colorlinks=true,citecolor=red,linkcolor=black]{hyperref}
\NameTag{AuthorOneLastName, \today}
\begin{document}

\nolinenumbers

\title{Future-proofing geotechnics workflows: accelerating problem-solving with large language models}

\author[1,2]{Stephen Wu}
\author[3]{Yu Otake}
\author[3]{Daijiro Mizutani}
\author[1]{Chang Liu}
\author[3]{Kotaro Asano}
\author[3]{Nana Sato}

\author[4]{Hidetoshi Baba}
\author[5]{Yusuke Fukunaga}
\author[4]{Yosuke Higo}
\author[3]{Akiyoshi Kamura}
\author[6]{Shinnosuke Kodama}
\author[3]{Masataka Metoki}
\author[3]{Tomoka Nakamura}
\author[3]{Yuto Nakazato}
\author[3]{Taiga Saito}
\author[7]{Akihiro Shioi}
\author[8]{Masahiro Takenobu}
\author[9]{Keigo Tsukioka}
\author[4]{Ryo Yoshikawa}

\affil[1]{Research Organization of Information and Systems, The Institute of Statistical Mathematics, Tokyo, Japan. Email: stewu@ism.ac.jp}
\affil[2]{The Graduate University for Advanced Studies, Department of Statistical Science, Tokyo, Japan.}
\affil[3]{Department of Civil and Environmental Engineering, Tohoku University, Miyagi, Japan.}
\affil[4]{Department of Urban Management, Graduate School of Engineering, Kyoto University, Kyoto, Japan.}
\affil[5]{Coastal Development Institute of Technology, Tokyo, Japan}
\affil[6]{Civil Engineer,Civil Engineering Group,Urban and Civil Project Department,NIKKEN SEKKEI LTD, Tokyo, Japan.}
\affil[7]{Disaster Prevention Solution Department, KOZO KEIKAKU ENGINEERING Inc., Tokyo, Japan}
\affil[8]{Port and Harbor Department, National Institute for Land and Infrastructure Management, Yokosuka, Japan}
\affil[9]{Railway Technical Research Institute, Tokyo, Japan.}

\maketitle

\begin{abstract}
The integration of Large Language Models (LLMs) like ChatGPT into the workflows of geotechnical engineering has a high potential to transform how the discipline approaches problem-solving and decision-making. This paper delves into the innovative application of LLMs in geotechnical engineering, as explored in a hands-on workshop held in Tokyo, Japan. The event brought together a diverse group of 20 participants, including students, researchers, and professionals from academia, industry, and government sectors, to investigate practical uses of LLMs in addressing specific geotechnical challenges. The workshop facilitated the creation of solutions for four different practical geotechnical problems as illustrative examples, culminating in the development of an academic paper. The paper discusses the potential of LLMs to transform geotechnical engineering practices, highlighting their proficiency in handling a range of tasks from basic data analysis to complex, multimodal problem-solving. It also addresses the challenges in implementing LLMs, particularly in achieving high precision and accuracy in specialized tasks, and underscores the need for expert oversight. The findings demonstrate LLMs' effectiveness in enhancing efficiency, data processing, and decision-making in geotechnical engineering, suggesting a paradigm shift towards more integrated, data-driven approaches in this field. This study not only showcases the potential of LLMs in a specific engineering domain, but also sets a precedent for their broader application in interdisciplinary research and practice, where the synergy of human expertise and artificial intelligence redefines the boundaries of problem-solving.
\end{abstract}

\section{Introduction}
Geotechnical engineering, a pivotal field within civil engineering, is concerned with the behavior of earth materials and the robustness of ground structures. It encompasses the study and practical application of soil and rock mechanics, geophysics, and other related sciences to the construction and maintenance of infrastructure. Despite the advances in computational methods and instrumentation, the field faces perennial challenges such as unpredicted subsurface conditions, environmental constraints, sustainability demands, and the increasing complexity of construction projects. Such challenges often result in costly overruns, safety risks, and extended timelines. The inherent variability of geotechnical properties and the localized nature of geological data add layers of uncertainty that traditional methods struggle to manage effectively. As a discipline that supports the foundation of infrastructural development, there is an exigent need for innovative solutions that can offer enhanced predictive capabilities, optimize design processes, and streamline decision-making. The integration of robust analytical tools that can handle complex, variable data with agility and precision is not just an option but a necessity for future-proofing the discipline. The importance of integrating machine learning and data science into geotechnics has gained higher attention in the recent years~\cite{PHOON2022967,PhoonZhang:2023}, yet the field is still struggling to take fully advantage of the rapid advancement of modern statistical tools~\cite{PHOON2022101189}.

Large language models (LLMs) represent a significant leap in the field of artificial intelligence (AI), particularly within the domain of natural language understanding and generation. These models are designed to comprehend and manipulate language with a near-human level of subtlety and nuance. Trained on extensive datasets encompassing a wide array of knowledge, LLMs represent a new wave of foundation models to generate, translate, summarize, and interpret text with human-like proficiency. Integrating with other analytical tools, LLMs can assist in data analysis, automate repetitive tasks, provide educational support, and facilitate research by synthesizing information from a broad corpus of knowledge. In fields where specialized knowledge is paramount, such as geotechnical engineering, LLMs have the potential to act as a powerful auxiliary tool. Kumar~\cite{GeoGPT:2024} has demonstrated some direct uses of LLMs in basic text-related tasks in geotechnical engineering. LLMs can process technical documents, extract relevant data, predict outcomes based on historical patterns, and even generate preliminary design concepts. Their ability to rapidly assimilate and apply complex instruction sets make them an ideal candidate for tasks that require both depth of knowledge and breadth of application. Furthermore, LLMs' iterative learning capacity means that they can continuously refine the outputs based on user feedback, leading to increasingly accurate and relevant applications over time. As industries gravitate towards digitalization and automation, LLMs stand out as a transformative asset that can augment human expertise and elevate the work style within geotechnical engineering.

This paper aims to explore the transformative potential of LLMs, such as ChatGPT~\cite{ChatGPT}, within geotechnical engineering. By aligning with the current and future needs of the field, the study seeks to establish a forward-thinking approach to workflow acceleration and problem-solving. The objectives are twofold: first, to assess the applicability and efficacy of LLMs in handling hands-on geotechnical tasks and second, to envision a paradigm where such AI-driven tools become integral to the geotechnical profession. The paper will delve into various case studies and practical examples from a recent workshop organized by the authors to demonstrate LLMs' role in simplifying complex problems and enhancing decision-making processes. The study intends to provide a comprehensive evaluation of its benefits and limitations. Ultimately, the goal is to pave the way for a new work style in geotechnical engineering that is more efficient, accurate, and adaptable to the rapidly evolving demands of the construction industry and environmental stewardship. This exploration is not only an academic exercise but also a critical step towards realizing the potential of AI in one of the most foundational realms of engineering.

\section{Background}
\subsection{Geotechnical Engineering: Current State-of-the-Art}
Geotechnical engineering is a field characterized by its diverse challenges, which arise primarily from the unpredictable nature of ground conditions. Typical problems include soil instability, unexpected subsurface materials, groundwater complications, slope failures, liquefaction, and foundation integrity issues. These challenges are critical as they directly impact the safety, sustainability, and financial viability of infrastructure projects. Solutions in geotechnical engineering are as varied as the problems, often requiring a combination of empirical methods, analytical models, and innovative technologies. Soil stabilization techniques, such as the use of geotextiles and earth reinforcement, are common responses to unstable grounds. Ground improvement methods like compaction, grouting, or the installation of piles are employed to enhance soil strength and mitigate settlement issues. In the case of water-related challenges, dewatering, waterproofing, and drainage systems are strategically deployed. Advanced geotechnical instrumentation and monitoring are also integral, providing real-time data that inform adaptive management strategies during and after construction.

The development and implementation of these solutions depend heavily on accurate site investigation and analysis. Geotechnical engineers rely on a suite of methods like drilling, sampling, in-situ testing, and geophysical surveys to gather subsurface data. This data is then analyzed using various software tools to predict behavior under different loading conditions and to design appropriate engineering responses. The evolution of computer-aided design (CAD) and the integration of Building Information Modeling (BIM) have significantly enhanced the capacity for precise modeling and simulation in geotechnical engineering. Moreover, sustainability considerations are increasingly driving the adoption of eco-friendly materials and techniques that minimize environmental impact while still providing structural integrity.

Geotechnical engineering workflows are complex and multidisciplinary, encompassing a cyclical process from initial assessment to post-construction evaluation. The workflow begins with a desktop study to collect geological, hydrological, and environmental data, followed by comprehensive field investigations and laboratory testing to determine ground characteristics and develop an initial conceptual site model. This model is essential for detailed design and analysis, utilizing various tools ranging from traditional calculations to advanced software for simulation and modeling. Collaboration is crucial in these projects, involving geologists, engineers, architects, and construction teams working together through project management workflows to share and effectively use information. As the project progresses, design and testing phases are iteratively conducted to refine solutions, adapt to new findings or changing conditions, and ensure compliance with design specifications and safety standards, underlining the importance of quality assurance and control throughout the project lifecycle.

In recent years, digital transformation has begun to reshape these traditional workflows. The adoption of digital tools and platforms enables more integrated and dynamic work processes. Data management systems are increasingly sophisticated, allowing for better handling of the increasing amounts of data in geotechnical projects. Additionally, emerging technologies such as AI and machine learning are beginning to be integrated, offering potential for more predictive and adaptive workflows that can enhance decision-making and risk management. 
As the field continues to evolve, it is expected that these technological advancements will further streamline geotechnical workflows, leading to greater efficiencies and better outcomes for engineering projects.

\subsection{Large Language Models and ChatGPT}
Large Language Models (LLMs) like GPT-3~\cite{NEURIPS2020_1457c0d6} or BERT~\cite{devlin-etal-2019-bert} have revolutionized the field of computational linguistics, bringing forth capabilities that extend far beyond basic text generation. These models, characterized by their deep learning architectures and trained on diverse internet text, have demonstrated proficiency in understanding and generating human-like text. LLMs are particularly adept at tasks requiring complex language understanding, such as answering questions, summarizing lengthy documents, translating languages, and even generating code. Their significance to interdisciplinary research in engineering cannot be overstated; they have the potential to accelerate research by quickly synthesizing literature, drafting research papers, and proposing hypotheses. Their impact has been seen across different disciplines, such as materials science~\cite{Zheng:2023aa,D3DD00113J} and geoscience~\cite{deng2023k2,Ma2023:GeoLLM}, but not yet in geotechnics.

In the realm of geotechnics, the capabilities of LLMs to assist in multimodal modeling and decision-making are invaluable. They can serve as a bridge between various disciplines, enabling engineers to integrate insights from environmental science, material science, and structural dynamics into cohesive solutions. By processing natural language queries, LLMs can rapidly search through vast databases of scientific literature and return concise summaries. They can also interpret and convert complex text data into more understandable formats, such as knowledge graphs, supporting engineers in making informed decisions. This becomes particularly relevant when dealing with the intricate data analyses often required in geotechnical engineering, where understanding the interplay between various physical and mechanical properties of soils and rocks is essential.

\begin{figure}
    \centering
    \includegraphics[width=0.8\linewidth]{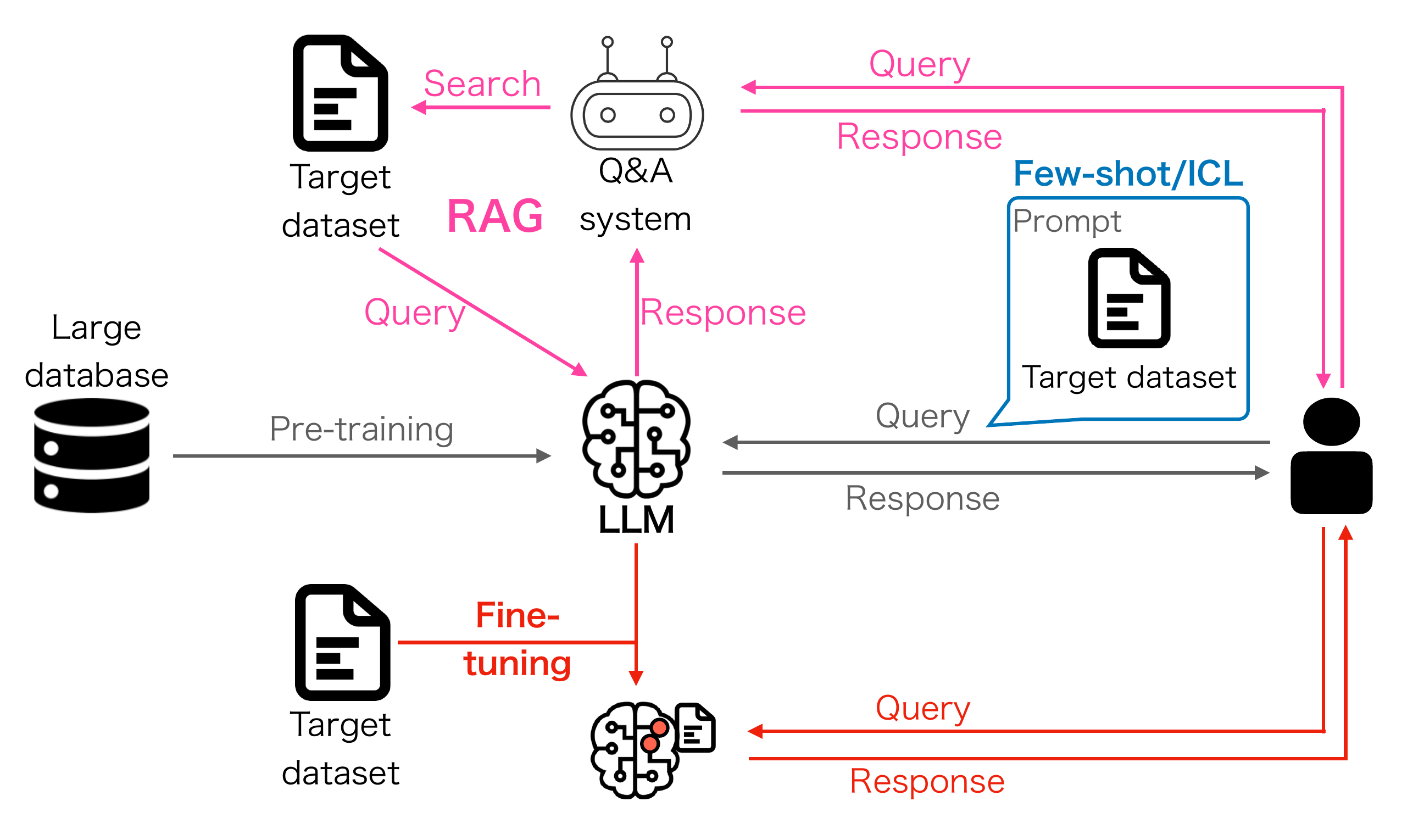}
    \caption{Overview of different extensions to LLM: (1) few-shot learning or in-context learning (ICL) refers to users directly providing examples to LLM, but does not alter the LLM model. (2) Fine-tuning refers to further training the LLM using a small set of data from the target task. (3) RAG refers to a technique to link the LLM with external databases to help LLM providing more accurate answers.}
    \label{fig:LLM}
\end{figure}

In addition, LLMs have advanced capabilities that extend beyond basic text generation, offering a range of sophisticated functionalities, such as text embedding, fine-tuning~\cite{dinh2022lift}, and Retriever-Augmented Generation (RAG)~\cite{NEURIPS2020_6b493230}, each enhancing their application in specialized domains (Figure~\ref{fig:LLM}). Text embedding involves transforming textual data into numerical vectors, capturing contextual nuances vital for tasks like sentiment analysis or providing the foundation for vector databases. Fine-tuning further specializes these models for specific tasks or domains by training them on targeted datasets, thereby improving their accuracy and relevance in particular contexts, such as legal language processing or specific scientific fields. RAG combines LLMs with a retrieval system, enabling the model to fetch relevant information from a vast corpus to inform its responses. This makes LLMs more informed and accurate, especially in fields requiring up-to-date knowledge. By leveraging these functionalities, LLMs can provide more nuanced, accurate, and contextually relevant outputs, significantly enhancing their utility in various sectors.

\section{Methodology}

A two-day workshop has been organized to explores the potential applications of LLMs within geotechnical engineering, with a diverse group of participants from academia, industry, and government sectors in Japan. The workshop facilitated a hands-on assessment of LLMs' capabilities, focusing on accelerating interdisciplinary research and simplifying complex analytical tasks through multimodal modeling. Because of the short time frame, we focused on a set of relatively simple techniques and tools related to LLMs. For example, few-shot learning (or in-context learning) was investigated rather than fine-tuning (Figure~\ref{fig:LLM}). The commercial implementation of RAG provided in ChatGPT~\cite{ChatGPT} by OpenAI was tested rather than using a customized RAG algorithm. Text embedding was tested using models and Application Programming Interface (API) systems provided by OpenAI. However, our preliminary results represent a useful starting point and serve as insightful guidelines for more advanced implementations of LLMs in geotechnical problems.

\subsection{Workshop Organization and Participants}
The workshop was held at The Institute of Statistical Mathematics in Tokyo, Japan, on November 3-4, 2023. The event brought together 20 diverse participants, including students, researchers, professors, engineers, and senior managers from academia, industry, and government sectors. The ambitious goal was to solve geotechnical problems using LLMs within the two-day workshop, culminating in the creation of an academic paper with LLM assistance within a month post-event. The composition of the participants was diverse, comprising 14 from academia (including 6 professors and 8 students), 3 from industry, and 3 from the government sector. The group included 2 data scientists and 18 geotechnical engineers, with varied experiences in LLMs and coding. The demographic spread was quite broad, with a notable skew towards male participants.

The workshop was meticulously organized by two data scientists and two geotechnical engineers, starting from early October 2023. Initially, 30 potential participants with various level of experience in machine learning and geotechnical engineering were shortlisted, resulting in 20 attendees for the physical workshop. To prepare participants, particularly those with limited LLM experience, two pre-event tutorial sessions were held, focusing on LLM basics and practical applications in geotechnical engineering. These sessions aimed to familiarize participants with LLMs, gather feedback, and identify specific geotechnical problems for exploration. Subsequently, participants were grouped into five working teams based on shared interests in specific geotechnical issues, combining senior experts with junior members having coding skills.

To maximize the workshop's effectiveness, the organizers focused on using ChatGPT as the primary tool due to its user-friendly, all-in-one service, as opposed to other LLM options. This decision was made considering the limited time and varying expertise levels of the participants. Although ChatGPT was the core tool, it was noted that most of the workshop's work could be replicated using other LLMs, including open-source or free alternatives, highlighting the adaptability and broad potential of LLMs in addressing geotechnical challenges.

During the workshop, each small group initiated their project by defining the problem setup and gathering relevant data. They then developed a preliminary workflow incorporating various functionalities of ChatGPT, which was shared with all participants for discussion and refinement. Post workflow finalization, the groups concentrated on solution implementation, supported by three all-around assistants. The workshop concluded with each group presenting their results and planning the academic paper to report their findings. Following the workshop, a significant update to ChatGPT enabled the construction of customized GPT models for specific tasks. Leveraging this, a smaller team of four participants was formed to develop GPT applications tailored to the different geotechnical challenges encountered during the workshop and to assist in composing the academic paper.

\subsection{Case Study Selection}
A total of four case studies were chosen for the workshop that were strategically aligned with both the challenges and opportunities in geotechnical engineering, particularly focusing on the application of LLMs like ChatGPT. These challenges include the scarcity of large and comprehensive databases in geotechnics, where data is often sparse, uncertain, and incomplete. Geotechnical data typically exists in various formats, including text and less organized numerical data, and is frequently multimodal, encompassing photos, textual descriptions, time series, and numerical figures. Building a set of relevant descriptors is one of the major bottlenecks to apply machine learning in geotechnical problems. The case studies were selected to demonstrate how LLMs could innovatively address these issues, taking advantage of different features of LLMs (Table~\ref{tab:case}). The success of these case studies underscores the potential of LLMs to transform geotechnical engineering, offering pathways to more efficient, accurate, and universally accessible solutions in this field.

\begin{table}
    \centering
\caption{Summary of the LLMs' features considered in each case study.}
\label{tab:case}
    \begin{tabular}{p{0.25\textwidth}|p{0.15\textwidth}|p{0.15\textwidth}|p{0.15\textwidth}|p{0.20\textwidth}}
        \hline
         Case study&  Multimodal data handling&  Python coding&  Knowledge extraction (RAG)& Text embedding\\
        \hline \hline
         Slope Stability Assessment&  Image + Text&  N/A&  N/A& Image descriptions\\
         \hline
         Microzoning by Seismic Risk&  Time series + Image + Text&  Data analysis&  Domain literature& N/A\\
         \hline
         Simulation Parameter Recommendation&  N/A&  N/A&  LIQCA manual& N/A\\
         \hline
         Site Similarity Prediction&  Numerics + Image + Text&  Data analysis + Plotting&  Python codes& N/A\\
         \hline
    \end{tabular}

\end{table}

\begin{itemize}
    \item \textbf{Slope Stability Assessment}: The motivation behind this study stemmed from the high cost and subjective nature of regular slope inspections in Japan, which are crucial for social safety. These inspections, typically conducted through site visits and visual assessments by experts, often lead to vague and subjective criteria, making it challenging to develop accurate quantitative models for risk assessment. This case study aimed to leverage the strong image recognition ability of deep learning and the multimodal capabilities of LLMs to create a virtual inspector model, potentially revolutionizing the way slope stability is assessed.
    \item \textbf{Microzoning by Seismic Risk}: The focus here was on the assessment of seismic risk for city planning and building code development, particularly in areas with high seismic activity. This case study aimed to test the ability of LLMs to integrate multimodal data sets, such as horizontal-vertical spectral ratio data~\cite{Yamada2017} and sensor locations, using natural language for decision-making. The challenge was in the subjective nature of seismic risk assessment, which often relies on the engineer’s judgment in combining spatial and spectral data.
    \item \textbf{Simulation Parameter Recommendation}: The motivation for this study came from the complexities involved in using LIQCA~\cite{Oka1994,Oka1999}, a simulation software used for effective stress dynamic analysis in geotechnical engineering~\cite{LIQCAtext}. Setting the thirteen necessary parameters, which include both mechanical and fitting parameters to simulate complex liquefaction behavior, requires specialized knowledge~\cite{Otake2022}. This case study explored the capacity of GPT models to provide recommendations for setting these parameters, potentially simplifying a complex and expert-driven process.
    \item \textbf{Site Similarity Prediction (Student Competition)}: This case was inspired by a challenge faced during the Joint Workshop on Future of Machine Learning in Geotechnics and Use of Urban Geoinformation for Geotechnical Practice in 2023~\cite{FOMLG2023}. The challenge involved the quantitative identification and extraction of geotechnically similar sites from a pre-documented database~\cite{Sharma2022}, a process known as the “site recognition challenge.” This approach is particularly appealing to geotechnical engineers as it allows for an experiential and judgment-based acceptance or rejection of identified similar sites, thereby making the process more explainable and relatable.
\end{itemize}

Each case study was chosen for its potential to demonstrate the versatility and utility of LLMs in addressing complex, multifaceted problems in geotechnical engineering, thereby offering insights into the future of problem-solving in the field.

\section{Results}
\subsection{Overview}
The results from the workshop's case studies collectively showcased the proficiency of LLMs, particularly ChatGPT, in geotechnical engineering tasks, revealing common strengths and limitations across diverse applications. In tasks like slope stability assessment and microzoning for seismic risk, the GPT models demonstrated a strong ability for multimodal data interpretation, effectively combining textual descriptions with image analysis to assess landslide risks and seismic data. This capability was further highlighted in the site similarity prediction, where GPT efficiently handled fuzzy classification problems, streamlining data analysis even for those without deep expertise in data science.

A recurring theme in these studies was the model's reliance on context and pre-provided examples for accurate predictions. For instance, in the simulation parameter recommendation, GPT required a standard parameter set to make logical descriptions and parameter predictions, reflecting its need for contextual grounding. Across all cases, while GPT showed potential in integrating complex data and providing insightful analyses, its output often required refinement to meet expert-level expectations. These findings underscore the promise of LLMs in geotechnical engineering, particularly in tasks involving multimodal data analysis and context-based learning, while also highlighting the necessity for tailored inputs and expert oversight for optimal results. In addition, GPT is a generative model, for which its output is uncertain in nature. On one hand, such feature can be used as a natural tool for uncertainty quantification. However, it also creates extra challenge for the tasks that require making precise and deterministic predictions.

\begin{figure}
    \centering
    \includegraphics[width=0.8\linewidth]{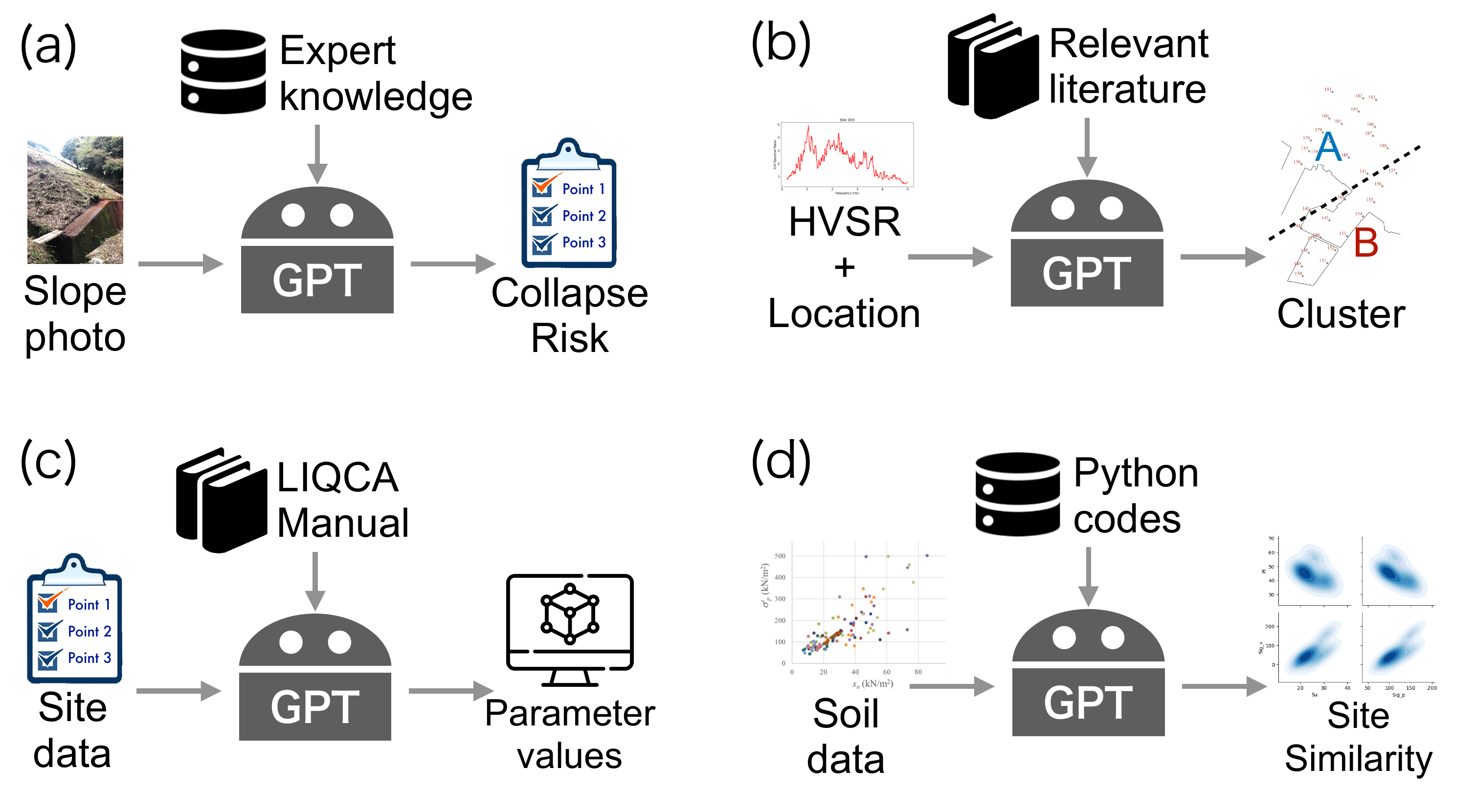}
    \caption{Overview of GPT uses in (a) the slope stability assessment study, (b) the microzoning by seismic risk study, (c) the simulation parameter recommendation study, and (d) the site similarity prediction study.}
    \label{fig:GPT}
\end{figure}

In the following, we will provide summaries of the results for each case study. Figure~\ref{fig:GPT} provides an overview of how GPT models are used in each study. We also provide customized GPT models used in the case studies that are publicly available for testing and our chat history for the studies (see Supplementary Information).

\subsection{Slope Stability Assessment}
In this study, an expert prepared ten diverse slope photographs from various regions in Japan, each labeled with a landslide risk percentage ranging from 0\% (very safe) to 100\% (already collapsed). The risk assessment criteria, recorded as text data, involved three key aspects: determining a baseline risk level based on soil moisture, adjusting this baseline with factors that influence slope stability (e.g., presence of retaining walls reducing risk), and setting the risk to 100\% if any slope cracks were observed. These criteria were designed to encapsulate the multifaceted nature of slope stability risk assessment, considering both visual and environmental factors.

To address this task, two GPT models were developed. Users were required to upload a slope photo, after which the models provided descriptions of the slope environment, assessed landslide risk, and assigned a risk percentage. The "Slope Stability Inspector: Zero-shot" model offered minimal preloaded knowledge for standard response generation, while the "Slope Stability Inspector: Few-shot" model included expert judgment and scoring scheme knowledge for landslide risk, supplemented with training examples for few-shot learning. The ten photos were divided into four for training and six for testing, with evaluations conducted thrice for each model to account for the models' memory of input history.

\begin{figure}
    \centering
    \includegraphics[width=0.8\linewidth]{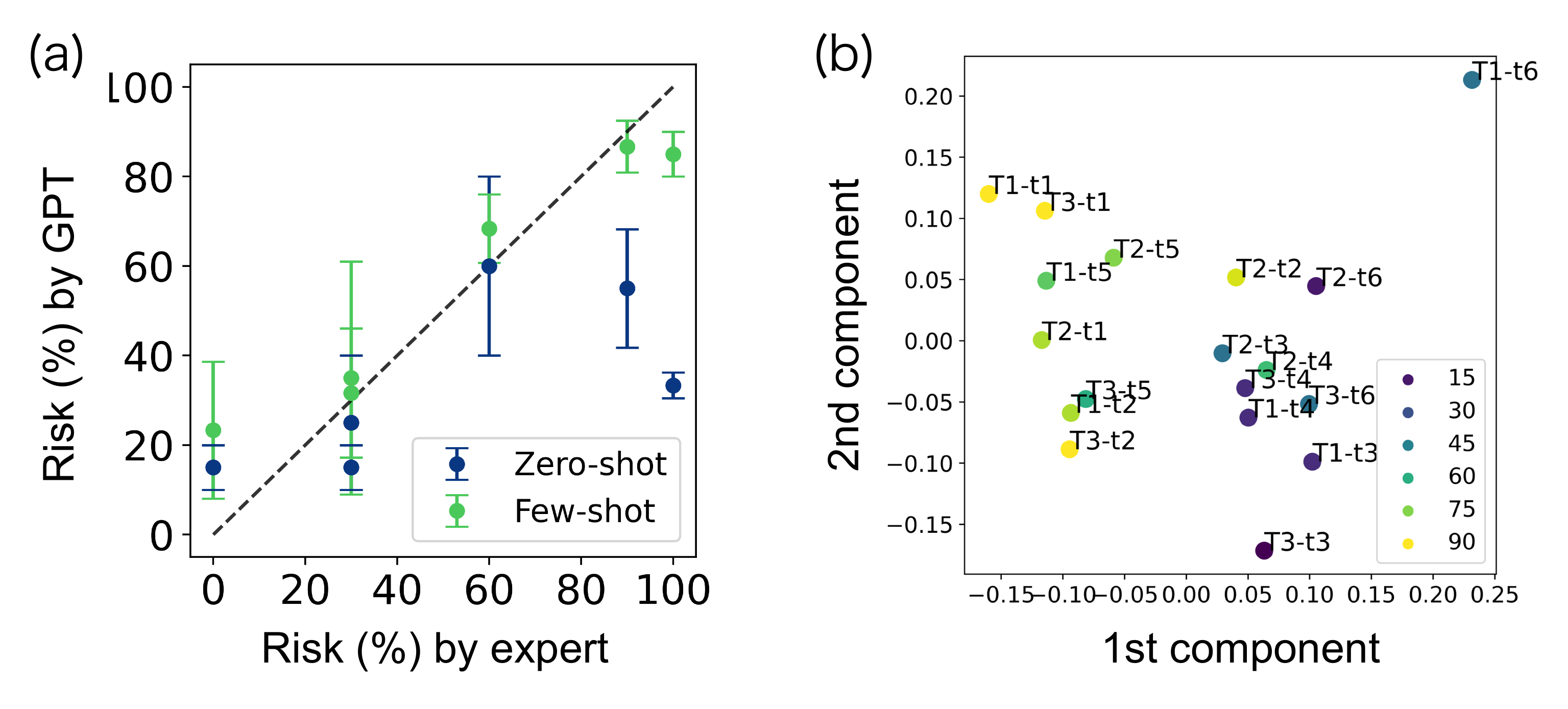}
    \caption{Results of collapse probability prediction from GPT models. (a) Prediction performance of the zero-shot and few-shot GPT models based on the expert-assigned risk values for the six test photos. The error bars denote the standard deviation of the GPT predictions over three independent trials. (b) First two principal components of the embedding vector for the photo descriptions generated by the few-shot GPT model. A total of 18 descriptions (3 trials of the 6 test photos) are plotted and the colors correspond to the predicted collapse probability.}
    \label{fig:slope}
\end{figure}

The study's results revealed the zero-shot GPT's capability in accurately describing photos relevant to landslide risk assessment, demonstrating strong multimodal modeling skills (see chat histories in Supplementary Information). However, its risk percentage predictions varied significantly. Conversely, the few-shot model showed more stable and expert-aligned risk predictions, highlighting GPT's ability in in-context learning and adjustment with minimal examples (Figure~\ref{fig:slope}a). To verify the descriptions generated by GPT are actually used for predicting the risk, we investigated the predictive power of the text embedding of the descriptions. Our analysis of text embeddings from GPT's descriptions showed a clear correlation between predicted collapse probabilities and principal components of the embedding vectors (Figure~\ref{fig:slope}b). These outcomes underscore GPT's unique strengths in handling vague concepts and reasoning, distinguishing it from conventional machine learning models. This study not only affirms GPT's potential in geotechnical engineering but also opens new avenues for embedding complex concepts into numerical spaces for machine learning applications.

\subsection{Microzoning by Seismic Risk}
In this study, data was collected and processed in the form of horizontal-vertical spectral ratio (HVSR) from various sites in Japan. This data was converted into PNG images, capturing the HVSR shapes, and was accompanied by the geographical locations of the corresponding sensors. The data set also included expert recommendations on how to group these sensors to form distinct sub-regions. This setup aimed to utilize the capabilities of LLMs to analyze and interpret complex seismic data, which is crucial for effective city planning and building code development in seismic risk areas.

\begin{figure}
    \centering
    \includegraphics[width=0.8\linewidth]{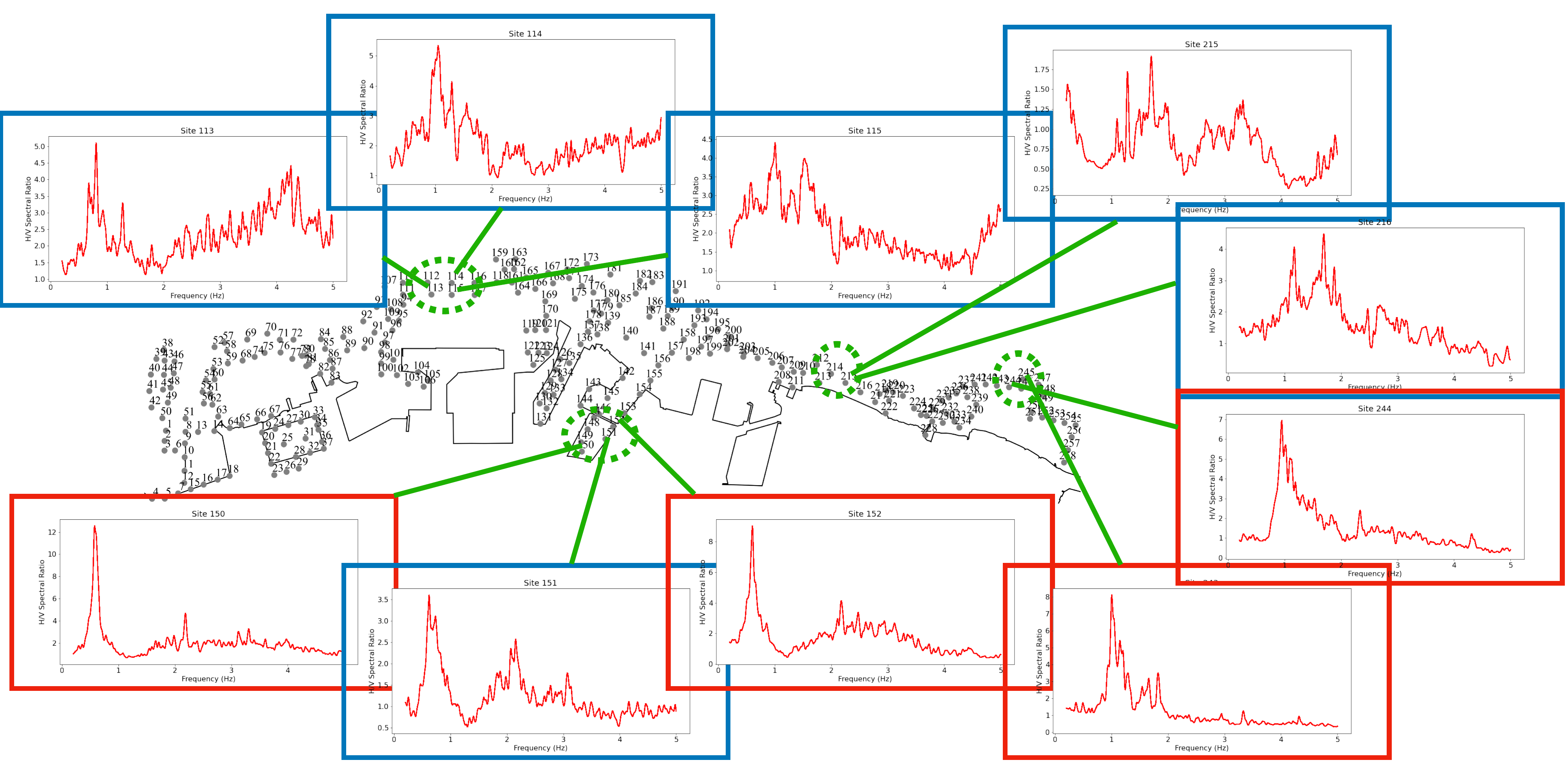}
    \caption{Results of microzoning predicted by the GPT model without domain knowledge based on HVSR curves at 10 selected sites from 4 location clusters. Four locations with a single outstanding peak below 2Hz are grouped together (red) and the other six locations are grouped together (blue), which is a reasonable result without considering the spatial distance between the sites.}
    \label{fig:micro1}
\end{figure}

To address this challenge, two GPT models were developed, each requiring users to upload figures of HVSR data. The first model was designed with minimal instructions, while the second model had access to prior knowledge concealed within a PDF document. The task was to assess the models' ability to describe the HVSR data and group sites accordingly. The study explored the difference in the models' performance with and without the document and further incorporated location data to assess the impact of spatial information on the grouping task. The models were tasked to mimic the expert's criterion that clusters HVSR curves with similar shapes together. However, sites that are close to each other should be clusters into the same group. Limited by the maximum number of files to be uploaded to ChatGPT at once, we selected 10 sites with a mixture of HVSR shapes obtained from four sensor clusters (Figure~\ref{fig:micro1}).

\begin{figure}
    \centering
    \includegraphics[width=0.8\linewidth]{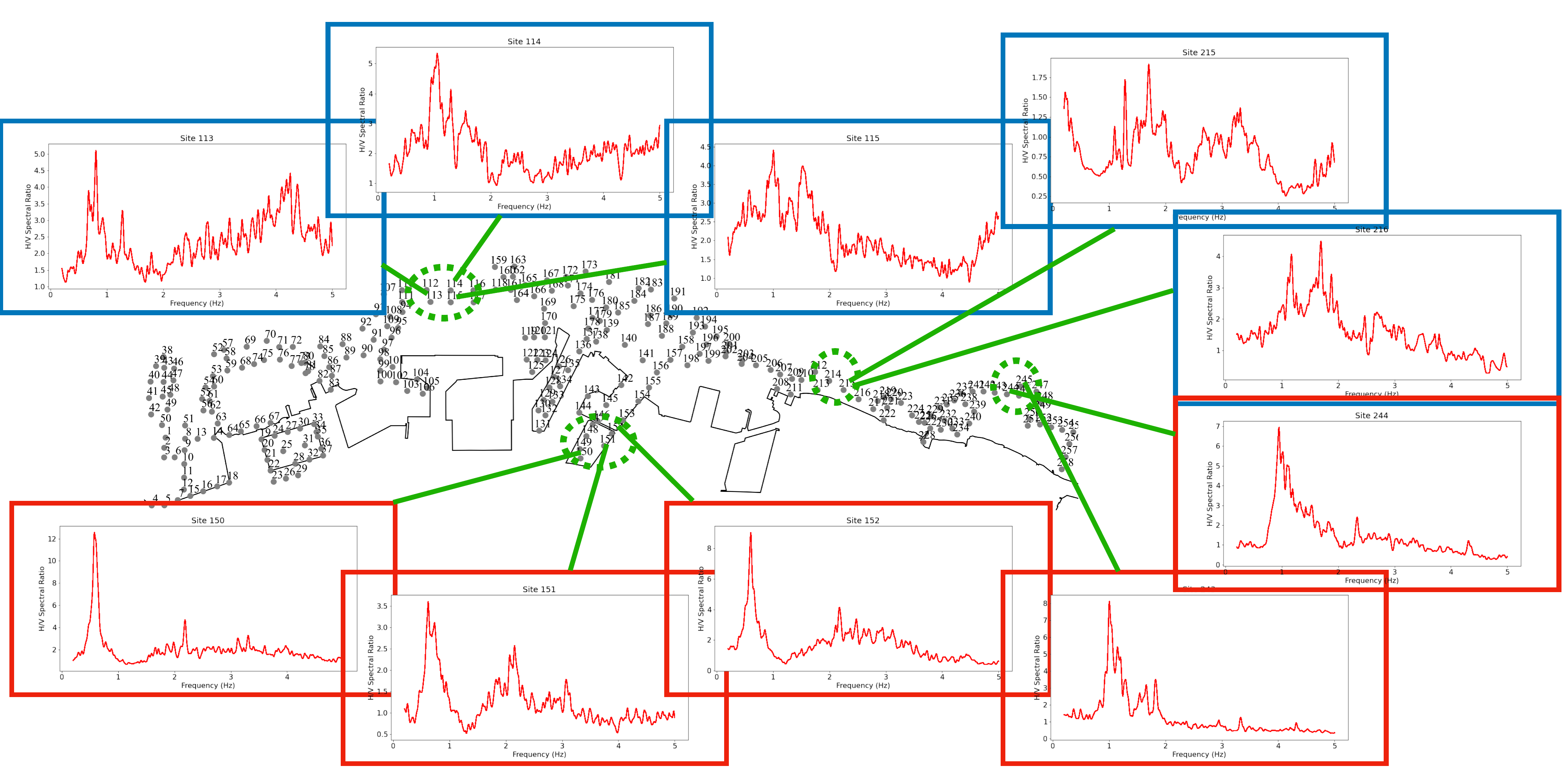}
    \caption{Results of microzoning predicted by the GPT model with domain knowledge based on HVSR curves at 10 selected sites from 4 location clusters. The result is almost the same as Figure~\ref{fig:micro1}, except one of the blue site is now assigned to be red because the neighboring sites are all grouped as red.}
    \label{fig:micro2}
\end{figure}

The results indicated that the baseline GPT model could identify key features in the HVSR data, such as the rough location and width of peaks, particularly those significant in the 0.2Hz-2Hz range (see chat histories in Supplementary Information). When location data was provided, the model could generate Python codes to decipher spatial relationships between sites, leading to grouping results that aligned precisely with expert recommendations (Figure~\ref{fig:micro1} and~\ref{fig:micro2}). However, it was noted that the performance in peak recognition and HVSR shape analysis was limited, attributed to the specialized nature of GPT's image model for more complex photos. Despite these limitations, the study demonstrated LLMs' potential in leveraging multimodal data for decision-making in geotechnical engineering, suggesting avenues for improvement, such as few-shot learning and the incorporation of analysis codes as prior knowledge.

\subsection{Simulation Parameter Recommendation}
\begin{figure}
    \centering
    \includegraphics[width=0.6\linewidth]{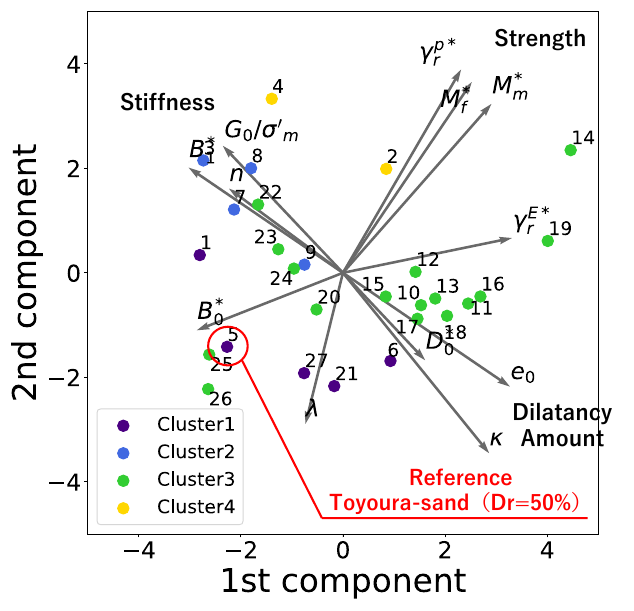}
    \caption{First two principal component space of the 13 dimensional parameter space of LIQCA. A generally understandable feature space was found, indicating stiffness, strength, and dilatancy level in three different directions. The different colors represent four soil categories grouped by geotechnical knowledge: (1) Cluster1 --- experimental sand (loosely packed, Dr=50\%-70\%) usually has a relatively high stiffness and dilatancy but strength is low. (2) Cluster2 --- fill soil (reclaimed soil, such as in Port Island, Rokko Island, etc.) typically has a high stiffness but low dilatancy. (3) Cluster3 --- natural deposit soil (relatively high fine particle content rate) has very diverse properties, scattering around the whole space. (4) Cluster4
    --- experimental sand (densely packed, Dr=75\%) usually has a distinctively high stiffness and strength.}
    \label{fig:LIQCA1}
\end{figure}

In this study, 27 high-quality soils with well-documented parameter values from the literature were utilized~\cite{LIQCAtext,Otake2022,Otsushi2010437}. Principal component analysis (PCA) was performed on this dataset, with the first two principal components used to visualize the high-dimensional parameter space (Figure~\ref{fig:LIQCA1}). This approach was independent of the GPT model's construction, with the information about these soils not provided to the model. Instead, the reduced dimension space served as a basis for evaluating the GPT model's performance. The complexity of dealing with 13 parameters was mitigated by mapping these values into a simpler two-dimensional space, facilitating a more effective assessment of the model’s predictive performance.

The problem setup is divided into two parts. First, from the 27 soils collected, we provided GPT model with an example of LIQCA parameters for Toyoura sand (relative density Dr=50\%), a world-renowned experimental sand, as a reference point. GPT model was then instructed to estimate the LIQCA parameters for Toyoura sand, characterized as densely-packed with a high relative density (Dr=75\%). Additionally, we assessed GPT's knowledge of the general soil types, including sand and reclaimed soil, in Niigata City, an area well-known for liquefaction. The intention is to determine if GPT can contribute to the estimation of LIQCA parameters using its extensive prior knowledge of sand.

The results showed that the GPT model could make logical descriptions of the sand samples and pick up relevant context from the provided manual. However, it struggled to provide numerical values for the parameters without a standard parameter set as a reference. The model understood terms like "compaction" and "high dilatancy," predicting numerical values for LIQCA parameters, but its adjustments didn't entirely align with expert expectations. Despite this, the model showed a reasonable understanding of different sands, such as Niigata sand, and was able to distinguish between different geotechnical properties. The model tended to prioritize general geotechnical knowledge over specific manual information, indicating a bias towards more broadly available literature. This trial highlighted the GPT model's potential in geotechnical engineering, paving the way for further fine-tuning and integration into tools like LIQCA.

\begin{figure}
    \centering
    \includegraphics[width=1.0\linewidth]{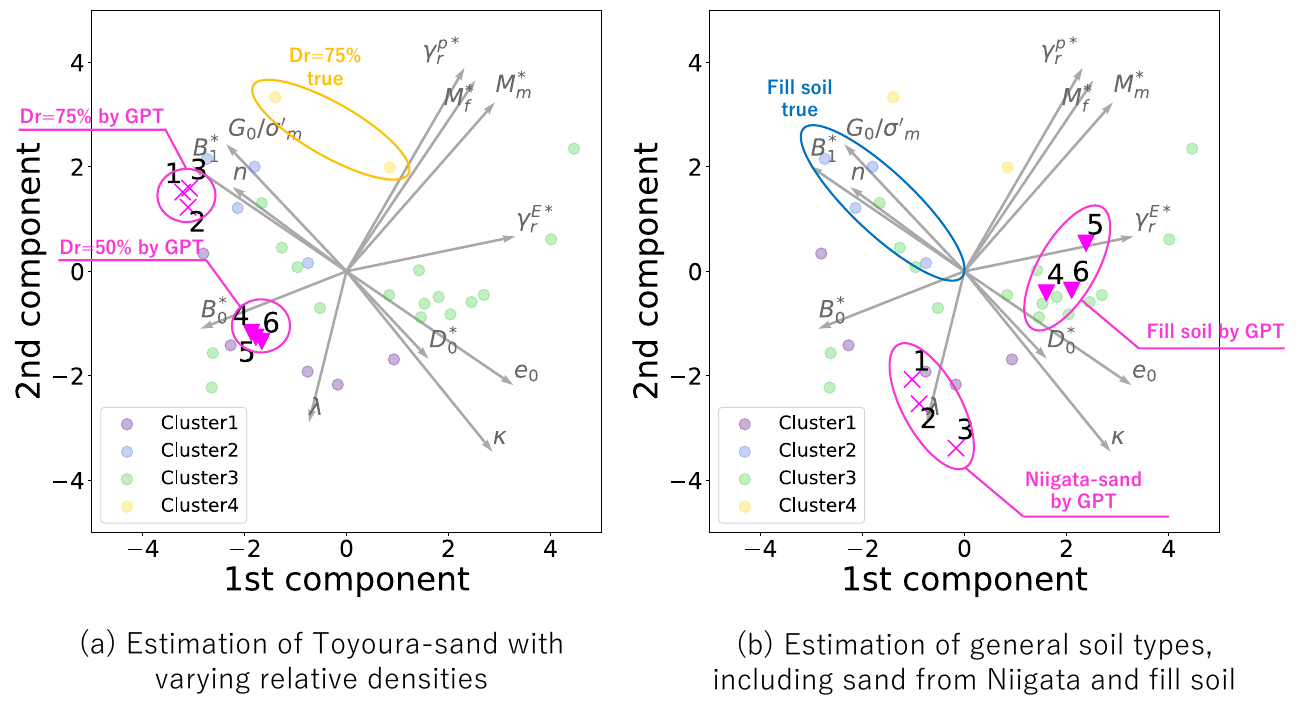}
    \caption{Recommeded parameter values by the GPT models projected to the same space as in Figure~\ref{fig:LIQCA1}. (a) Estimation of Toyoura-sand with varying relative densities, (b) Estimation of general soil types, including sand from Niigata and fill soil. It is evident that GPT model primarily expresses the "difference in relative density" by adjusting parameters related to stiffness. The parameters for the sand from Niigata City, known for its homogeneous grain size, are estimated to be similar to those of Toyoura sand with a relative density (Dr) of 50\%. The fill soil is differentiated from both Niigata-sand and Toyoura sand mainly by altering the parameters associated with strength.See chat histories in Supplementary Information for more details.}
    \label{fig:LIQCA2}
\end{figure}

\subsection{Site Similarity Prediction (Student Competition)}
In the site recognition problem tackled during the student contest, the data used comprised clay properties collected from 16 different sites globally, along with a separate dataset containing similar properties from an unknown target site. Each site was defined as a collection of records from tests conducted within a project boundary, with data including undrained shear strength, preconsolidation stress, corrected cone tip resistance, plastic index, and vertical effective stress. This comprehensive dataset provided a robust foundation for assessing site similarity based on these key geotechnical properties.

The problem setup involved creating scatter plots for 10 selected property pairs from the 5 clay properties for each of the 16 sites, using a standardized x-range and y-range for contour consistency. Python functions for generating these plots were provided to the GPT model as prior knowledge to ensure uniformity in plot creation. The same set of scatter plots was then created for the unknown target site. Each site’s scatter plots were summarized in a single image with a two-by-five layout, which was then used by the GPT model to compare image similarities between the target site and the other 16 sites. The comparison was based solely on the shape contours of the 10 plots in these images, providing a visual similarity measure.

\begin{figure}
    \centering
    \includegraphics[width=0.6\linewidth]{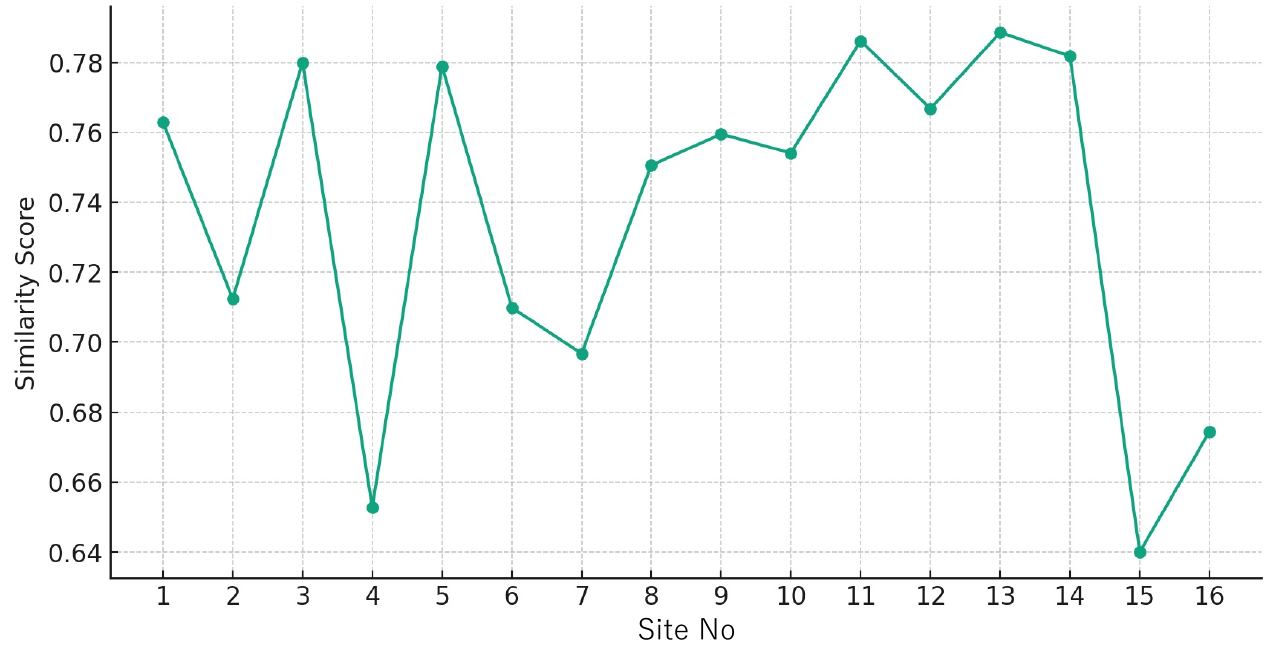}
    \caption{Result of similarity measure between the unknown target site and the 16 sites from the global database. The unknown site is supposed to be similar to site 13, which has the highest similarity value given by the GPT model.}
    \label{fig:TOC304}
\end{figure}

The results were promising, with the GPT model ranking the 16 reference sites according to similarity in a manner generally consistent with those obtained through a hierarchical Bayesian site similarity measure~\cite{Sharma2022,Ching202104021069} (Figure~\ref{fig:TOC304}, Table~\ref{fig:TOC304}). GPT successfully discriminated image features and quantified site-specific similarities without relying on explicit mathematical formulas. This study highlighted GPT’s capability as a convenient tool for accelerating basic data analysis tasks, particularly beneficial for individuals without expertise in data science. Additionally, it served as a proof-of-concept for GPT's potential in handling fuzzy classification problems through a combination of image and natural language analysis, marking a significant advancement in geotechnical data processing.

\begin{table}
    \centering
\caption{Comparison of Top 7 Similarity Ranking Results.}
\label{tab:similarity}
    \begin{tabular}{p{0.10\textwidth}|p{0.10\textwidth}|p{0.10\textwidth}}
        \hline
        Rank & HBM & GPT model \\ \hline \hline
        1st&Site 13&Site 13 \\ \hline
        2nd&Site 11&Site 11 \\ \hline
        3rd&Site 14&Site 5 \\ \hline
        4th&Site 10&Site 14 \\ \hline
        5th&Site 3&Site 3 \\ \hline
        6th&Site 9&Site 1  \\ \hline
        7th&Site 5&Site12 \\ \hline
    \end{tabular}
\end{table}

\section{Discussion}
\subsection{Potential of LLMs in geotechnics}
\begin{figure}
    \centering
    \includegraphics[width=0.8\linewidth]{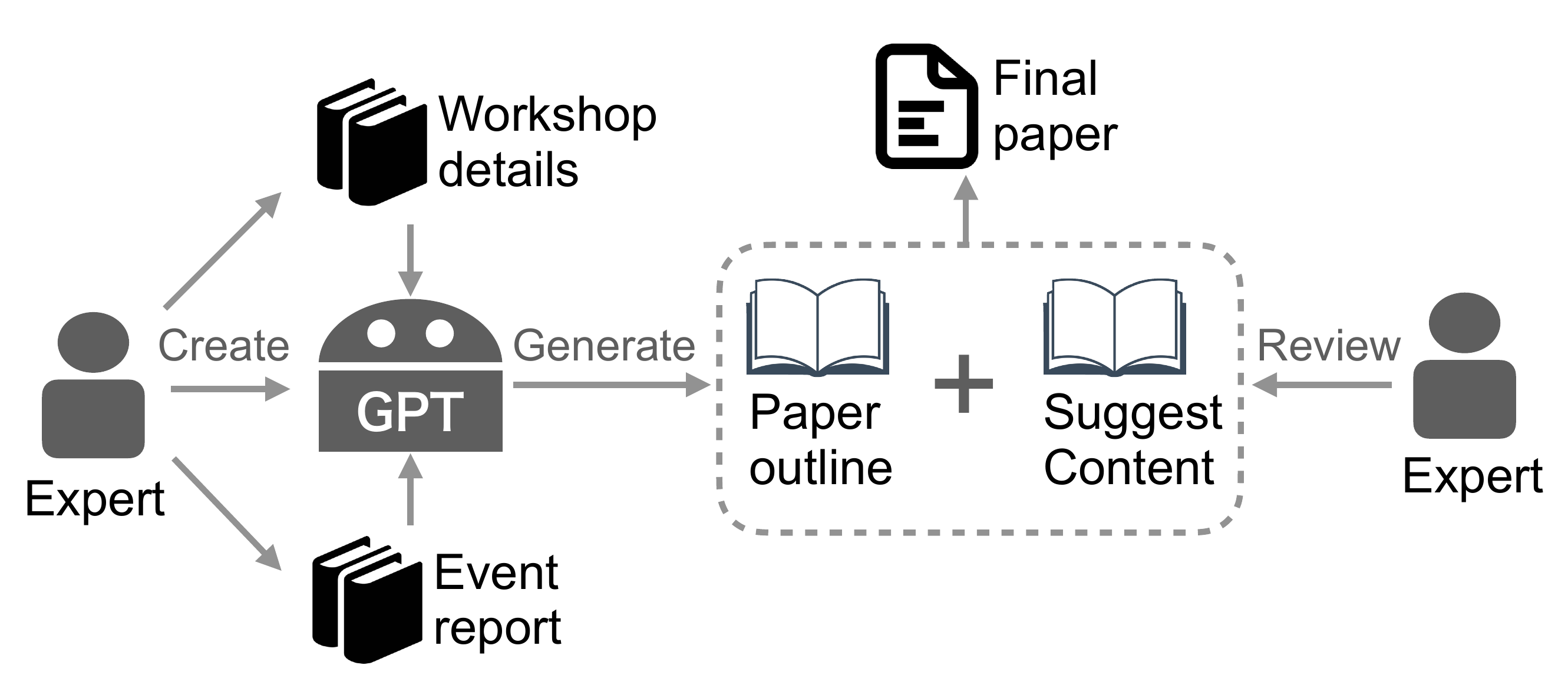}
    \caption{Workflow of the production of this paper. The authors are responsible for the production of the documents recording the workshop details and results of the studies conducted during the event, building the customized GPT model, and the final review and composition of the paper. The GPT model serves as a generator of writing idea, including the overall structure of the paper, paper title, and recommending corresponding content in each section of the paper.}
    \label{fig:paper}
\end{figure}

The potential of LLMs in geotechnical engineering is highlighted by their proficiency in handling a wide range of routine tasks that are common across various fields. LLMs excel in summarizing documents, translating text, conducting basic data analysis, and managing simple image-based design tasks. Such abilities are particularly beneficial in geotechnical engineering, where the processing and interpretation of diverse data forms are frequent necessities. For instance, LLM plays an important role in the creation of this paper. We built a customized GPT model, which is also publicly available for testing (see GPT links in Supplementary Information), to solely support the writing of this paper. The details of our workshop organization and schedule, as well as a detail report on all the results of the event were provided to the GPT model as prior knowledge. Then, the model was used to provide recommendations on the title of the paper, paper structure, and contents of each section (Figure~\ref{fig:paper}). However, the potential of LLMs extends beyond these general features.

In this paper, we demonstrated that by building on this foundational utility, LLMs' capacity for natural language processing allows them to serve as a pivotal tool for multimodal data analysis, a common requirement in geotechnical engineering. This becomes crucial considering the often sparse nature of geotechnical data, which typically includes a mix of text, images, time series, and numerical figures. For example, LLMs can interpret soil report narratives alongside numerical test results, offering a comprehensive understanding of geotechnical conditions. Furthermore, their ability to navigate and utilize the inherent ambiguities and subtleties of human language makes them adept at leveraging vague information, which is frequently encountered in geotechnical data due to the lack of quantitative details. Through text embedding, such vague information can be converted to numerical vectors that provide extra source of input data for training machine learning models.

The generative nature of LLMs also positions them effectively for uncertainty quantification and handling missing data, enhancing their applicability in geotechnical engineering where such issues are prevalent. They can generate predictive models or extrapolate from existing data to fill gaps, aiding in more accurate risk assessments or design decisions. Moreover, LLMs' capability to provide reasoned predictions is a crucial advantage over conventional machine learning methods, especially important for geotechnical engineers who often face complex decision-making scenarios. This reasoning ability, combined with the versatility and multimodal data processing capacity of LLMs, suggests that they could be more effective than traditional machine learning methods in addressing the unique challenges of geotechnical engineering. The integration of LLMs into geotechnical workflows, as demonstrated in various applications, signifies a shift towards more efficient, insightful, and data-driven approaches in the field.

Last but not least, the interdisciplinary nature of geotechnical engineering, often involving a blend of soil mechanics, environmental considerations, and structural dynamics, makes LLMs particularly valuable. Our workshop showcased how participants (mainly beginners in LLMs and data science), within just two days, could leverage LLMs to bridge the gap between geotechnics and data science. This rapid integration into interdisciplinary research is indicative of LLMs' ability to streamline complex, cross-disciplinary studies.

\subsection{Challenges of LLMs in geotechnics}
While Large Language Models (LLMs) offer significant potential in geotechnical engineering, there are inherent challenges in their application to complex engineering tasks. A fundamental limitation arises from the very nature of LLMs: they generate responses based on a probabilistic model that predicts the next word in a sequence. The "logic" presented in their answers is not true logical reasoning but rather an illusion created by the model's ability to capture probabilistic relationships between words. This means that while LLMs can provide seemingly logical responses, they should not be solely relied upon for final solutions in complex engineering tasks. For example, when tasked with predicting the stability of a slope, an LLM might generate a coherent and logical-sounding response, but this answer is based on word probabilities rather than a deep understanding of geotechnical principles.

Another challenge lies in the LLMs' ability to embed text-based data into numerical vectors while retaining the context. Although this feature enhances the efficiency of descriptor extraction from textual data, precisely controlling the focus of the context within these embeddings is difficult. In geotechnical engineering, where specific details can be crucial, such as in soil analysis reports, the inability to direct focus on a few keywords relevant to specific soil properties accurately can lead to incomplete or misleading interpretations.

The generative nature of LLMs also presents challenges, particularly when deterministic answers are required. LLMs are inherently designed to generate content, which can be problematic in scenarios demanding precise, unambiguous responses. As a result, additional tools and methods are often necessary to complement LLMs for such tasks. For instance, in generating specific design parameters for a geotechnical structure, an LLM might need to be paired with specialized software that can provide the required numerical precision.

Fine-tuning LLMs to produce highly accurate answers is another area of difficulty. While fine-tuning is expected to be an important technique to create LLM specialized in a unique task, the complexity of their training and the vastness of their data sources mean that achieving precision in their outputs, especially in a specialized field like geotechnical engineering, can be a nontrivial task. This is compounded when dealing with tasks that require high-precision numerical answers, where even the integration of external tools might not fully compensate for the LLMs' limitations. Effective interface design becomes crucial in such scenarios to ensure that the LLM can interact appropriately with other systems and tools, thereby enhancing the overall accuracy and utility of the outputs. For example, in applications requiring exact numerical data for soil properties, the interface between the LLM and geotechnical databases must be meticulously designed to facilitate precise data retrieval and interpretation.

\section{Conclusions}
In conclusion, the workshop and subsequent explorations into Large Language Models (LLMs) like ChatGPT have demonstrated significant potential for transforming geotechnical engineering workflows. The diverse applications and case studies discussed in this paper reveal LLMs as powerful tools capable of handling a wide range of tasks, from basic data analysis to complex, multimodal problem-solving. The ability of LLMs to process and interpret mixed forms of data, including text and images, and their capacity for natural language understanding, make them particularly valuable in the geotechnical field, where data is often sparse and multifaceted. These capabilities, combined with their generative nature, position LLMs as effective solutions for uncertainty quantification and handling incomplete data, common challenges in geotechnical projects.

However, the application of LLMs in this domain is not without its challenges. The nature of LLMs as probabilistic models that generate responses based on word relationships means they cannot be solely relied upon for final solutions in complex engineering tasks. The need for expert oversight and tailored inputs to achieve optimal results is evident. Additionally, fine-tuning LLMs for highly accurate and specialized responses in geotechnical engineering remains a complex task, highlighting the importance of effective interface design to ensure seamless integration with other systems and tools.

Looking ahead, the integration of LLMs into geotechnical engineering signifies a shift towards more efficient, data-driven, and insightful approaches. The interdisciplinary nature of geotechnical engineering, encompassing soil mechanics, environmental considerations, and structural dynamics, makes LLMs particularly valuable. The workshop exemplified how even beginners in LLMs and data science could rapidly integrate these tools into their work, underscoring the potential of LLMs to streamline complex cross-disciplinary studies. As the field evolves, LLMs are poised to play a pivotal role in shaping the future of geotechnical engineering, driving innovation and enhancing the efficacy of workflows in this foundational engineering realm.

\section{Acknowledgments}
This research was supported by ``Strategic Research Projects'' grant from ROIS (Research Organization of Information and Systems). Wu, Otake, Mizutani, and Liu were the four organizers of the workshop and contributed equally. Asano and Sato provided extra help on customized GPT development post-event. All other authors are listed according to alphabetical order of their last names. The preparation of this paper was supported by a customized version of ChatGPT-4 on December 4, 2023.

\section{Supplementary Information}

Supplementary Information is prepared as a separate document.

%
%
\bibliography{GeoLLM_ref}

\end{document}